\documentclass{article}

\usepackage{arxiv}

\usepackage[utf8]{inputenc} 
\usepackage[T1]{fontenc}    
\usepackage{hyperref}       
\usepackage{url}            
\usepackage{booktabs}       
\usepackage{amsfonts}       
\usepackage{nicefrac}       
\usepackage{microtype}      
\usepackage{lipsum}		
\usepackage{graphicx}
\usepackage{natbib}
\usepackage{doi}
\usepackage{tikz}
\usetikzlibrary{shapes,arrows,decorations.pathreplacing}
\usepackage{verbatim}
\usepackage{amsmath}

\title{Auspex: Building Threat Modeling Tradecraft into an Artificial Intelligence-based Copilot}

\date{} 					

\author{Andrew Crossman\thanks{Authors listed alphabetically.} \\ JP Morgan Chase
    \And Andrew R. Plummer \\ JP Morgan Chase
    \And Chandra Sekharudu  \\ JP Morgan Chase
    \And Deepak Warrier \\ JP Morgan Chase
    \And  Mohammad Yekrangian \\ JP Morgan Chase
}

\begin{document}
\maketitle

\begin{abstract}
	We present Auspex - a threat modeling system built using a specialized collection of generative artificial intelligence-based methods that capture threat modeling tradecraft. This new approach, called \textbf{\emph{tradecraft prompting}}, centers on encoding the on-the-ground knowledge of threat modelers within the prompts that drive a generative AI-based threat modeling system.  Auspex employs tradecraft prompts in two processing stages.  The first stage centers on ingesting and processing system architecture information using prompts that encode threat modeling tradecraft knowledge pertaining to system decomposition and description.  The second stage centers on chaining the resulting system analysis through a collection of prompts that encode tradecraft knowledge on threat identification, classification, and mitigation.  The two-stage process yields a threat matrix for a system that specifies threat scenarios, threat types, information security categorizations and potential mitigations.  Auspex produces formalized threat model output in minutes, relative to the weeks or months a manual process takes.  More broadly, the focus on bespoke tradecraft prompting, as opposed to fine-tuning or agent-based add-ons, makes Auspex a lightweight, flexible, modular, and extensible foundational system capable of addressing the complexity, resource, and standardization limitations of both existing manual and automated threat modeling processes.  In this connection, we establish the baseline value of Auspex to threat modelers through an evaluation procedure based on feedback collected from cybersecurity subject matter experts measuring the quality and utility of threat models generated by Auspex on real banking systems.  We conclude with a discussion of system performance and plans for enhancements to Auspex.   
\end{abstract}


\section{Introduction}

Threat modeling processes began to take shape in the 1960s and '70s with the development and expansion of shared computing resources that spanned government, corporate, and civilian network infrastructure, and the reflexive need to protect sensitive systems from external intrusion.  During the 1980s, methods of threat analyses resulting in protection and mitigation implementation recommendations for guarding systems against attacks became key features in computing infrastructure design \citep[see][]{Barnard1988}. The scope of information security needs by the end of the 1990s wrought the advent of methodologies for threat analyses based on structural models of attack methods \citep{Amoroso1994, SalterEtAl1998, Schneier1999} as well as the codification of threat modeling frameworks (viz. OCTAVE, detailed in \citealt{AlbertsEtAl1999}) and typological models (viz. STRIDE, \citealt{KohnfelderGarg1999}).  The STRIDE model in particular (reviewed in \citealt{STRIDE2022}, and used later in this paper) centers on mapping identified system threats to threat types to facilitate mitigation recommendations. 

In the decades that followed, a number of additional threat modeling frameworks emerged for addressing different facets of security threats such as assets, attackers, value and stakeholders, or a combination of approaches.  The LINDDUN framework \citep{DengEtAl2011}, for example, is privacy-centric complement to the STRIDE model based  a catalog of privacy threat types applied in system threat identification.  In contrast to STRIDE, the PASTA framework \citep{VelezMorana2015} takes a business value risk-centric perspective and focuses on threat modeling from the vantage point of potential attackers.  Moreover, frameworks like VAST \citep{VAST2018} emphasize the practicalities of simple collaborative threat modeling activities based on visuals of data flow and process flow diagrams common to many threat modeling frameworks.

In addition to the frameworks, a host of threat modeling tools have been developed to assist in the threat modeling process, including Microsoft's Threat Modeling Tool, OWASP Threat Dragon, OWASP pytm, IriusRisk, ThreatModeler, Threat Composer, inter alia.  Since the advent of widely-available generative AI models over the last few years, a broad swath of newer tools based on generative AI have become available within the academic and industry domains  (see Section~\ref{sec:related}).

Distilling from the expansive background presented above, we take \emph{threat modeling} to be a process that involves identifying and enumerating potential threats to systems, analyzing system vulnerabilities, prioritizing countermeasures, and recommending safeguards and mitigations \citep[for overviews from the government, nonprofit, and private sectors, respectively, see][]{CMS2024, OWASP2025, Boyd2021}.  The outcome of a threat modeling process is a \emph{threat model} - which typically includes a description of the system being modeled, a list of potential threats to the system, and a corresponding list of mitigations for each identified threat \citep{OWASP2025}.  

Threat modeling is a necessary activity for ensuring the security of critical government, corporate, and civilian computing infrastructure spanning global to local levels.  Yet, even with the aforementioned collection of frameworks, models, and tool kits in hand, threat modeling of systems in practice remains time- and resource-intensive, taking weeks to months to complete even for systems whose vulnerabilities represent substantial immediate risk. 

\subsection{Contributions}

In this paper, we introduce Auspex - a generative AI-based copilot that simplifies, accelerates, and enhances the end-to-end threat modeling process.  Auspex takes as input a user-provided system representation, e.g., a textual description or system architecture diagram, along with a set of modeling parameters, and outputs a threat model of the given system (see Figure~\ref{fig:e2e}). 

Our first contribution is the design and development of Auspex as a lightweight, flexible, modular, and extensible generative AI system that maps system representations to threat models in a manner that captures threat modeling tradecraft knowledge (Section~\ref{sec:auspexform}).   The design is based on our creation of \textbf{\textit{tradecraft prompting}} - a human-driven complement to ExpertPrompting \citep{xu2023expertpromptinginstructinglargelanguage, long-etal-2024-multi-expert} that encodes the detailed tradecraft practices of threat modelers within prompts and prompt chains (described further throughout Section~\ref{sec:tech}).    

Our second contribution is a technical formulation of the threat modeling task, including point-of-departure formulations of system representations, threat models, and mappings between them (Sections~\ref{sec:taskform} and ~\ref{sec:techform}).  The current threat modeling landscape is complex and nuanced, encompassing a wide variety of methodologies, frameworks, and tool kits, as well as practices that are idiosyncratic to threat modeling groups internal to government organizations and industry teams.  Given the amount of variation, it is necessary to formally standardize the threat modeling task that we address with Auspex in this work, and make Auspex amenable to methods of evaluation.  

Our third contribution is an initial evaluation procedure for measuring system performance based on subject matter expert feedback for Auspex-generated threat models over real banking systems (Section~\ref{sec:eval}).  The evaluation crucially captures judgments by experts on the overall value of Auspex in actually carrying out the threat modeling process.  We conclude with a discussion of how Auspex addresses current limitations facing threat modeling practice, as well as plans for expansion of our work (Section~\ref{sec:discussion}).   

\tikzstyle{boundt} = [draw, rectangle, rounded corners,
    minimum height=3em, minimum width=4em,align=center]

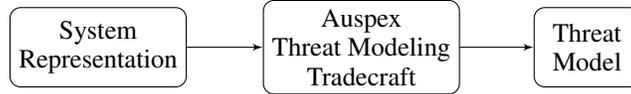
\begin{figure}
\centering
\begin{tikzpicture}[auto, node distance=2cm,>=latex']

\node (descS) [boundt] {System\\Representation};
\node (sys) [boundt, right of=descS, xshift=1.5cm] {Auspex\\Threat Modeling\\Tradecraft};
\node (tm) [boundt, right of=sys, xshift=1cm] {Threat\\Model};
\draw [->] (descS) to (sys);
\draw [->] (sys) to (tm);
\end{tikzpicture}
\caption{Auspex end-to-end system threat modeling in its simplest form.  Auspex takes in a system representation - an architecture diagram or a textual description - and outputs a threat model of the system.}
\label{fig:e2e}
\end{figure}

\subsection{Contrasting related work in AI-based threat modeling}
\label{sec:related}
As a preface to the formulation in the next section, we contextualize Auspex with respect to related AI-based efforts in the threat modeling domain.

Cyber Sentinel \citep{kaheh2023cybersentinelexploringconversational} aims to help security teams identify threats via LLM-supported processing of cybersecurity logs, events, and threat feeds, and then respond to those threats through LLM-generated security actions.  The underlying architecture is a task-oriented dialog system driven by GPT-4 and the use of chain-of-thought and self-consistency prompt engineering methods that propel engagement with users until the system is able to extract enough information to properly scope the user's security issue and action request.  The extracted information is used to derive possible security actions from a database of known controls and threat mitigations, which are then passed back to the user.  No formal evaluation of Cyber Sentinel is provided, and its broad scope differs from Auspex's focus on application-specific threat modeling.  

\cite{ElsharefEtAl2024} put forward a tool more focused on system threat modeling that aims to address the questions "what are we working on?" and "what can go wrong?" from the threat modeling manifesto \citep{tmm2025}.  Their AI-based threat modeling system is built from Llama 2-chat models supported by retrieval-augmented generation methods that draw from a corpus of system design documents and the National Vulnerability Database (NVD).  The system is given two tasks: help the user in understanding the system being threat modeled (i.e., answer the first question), and help the user with identifying potential security threats (i.e., answer the second question).  The RAG methods draw from system design documents to help with the former, and the NVD to help with the latter.  The evaluation method used 12 pdf documents containing descriptions of systems to be threat modeled.  Three questions were asked for each of the two tasks for each system, and the AI-generated output was judged by human annotators as satisfactory or not.  Formal results of the evaluation are not presented. 
 Note that the system differs from Auspex in the use of RAG and the lack of structure in both the threat modeling task and the generated threat output. 

STRIDE GPT \citep{STRIDEGPT2024} is an AI-powered tool specifically built for system threat modeling.  The tool is based on generative AI model families available through OpenAI, Anthropic, as well as the Gemini and Mistral model families.  STRIDE GPT ingests system diagrams, github repos, application descriptions, inter alia, and outputs a threat model that includes threats along with their STRIDE threat types, potential impacts, and suggested mitigation strategies.  The prompting methods used by STRIDE GPT are brief textual expert identity and task descriptions, while Auspex relies on tradecraft prompting that incorporates detailed information about threat modeling practice.  Since STRIDE GPT is open-source, a comparative evaluation of its generated threat models with Auspex-generated threat matrices is ongoing for later publication.    

PILLAR \citep{mollaeefar2024pillaraipoweredprivacythreat} focuses on system threat modeling using the privacy-centered LINDDUN framework.  PILLAR is also based on model families available through OpenAI as well as the Gemini and Mistral model families.  The input is a user-provided textual description of a system and/or its data flow diagram.  After input, there are three threat modeling options.  The first option is a zero-shot threat model generation using just the user input and the selected LLM to produce a generic baseline threat model centered on the LINDDUN threat categories.  The second option uses a single LLM agent that ingests the system description and a set of potential threats and determines if each threat is applicable to the system, while also providing an explanation for the decision.  The third option is a multi-agent approach, wherein different agents are given different points of view for addressing the relevance of threats to the system.  The agents engage in several rounds of analysis before a judge LLM uses their analyses to decide if a threat is relevant to the system.  Once a list of relevant threats is identified, an LLM receives each threat in turn, together with the system description, and a description of a collection of privacy patterns, and  outputs a list of relevant privacy patterns to mitigate each threat.  No formal evaluation of PILLAR is provided, though the open-source version is available for inspection.  The prompting methods are again brief textual expert identity and task descriptions, which contrast with Auspex's tradecraft prompting.  Moreover, Auspex's detailed tradecraft prompting allows for outputting high-quality threat matrices without resorting to agent-based add-ons. 

ThreatModeling-LLM \citep{yang2024threatmodelingllmautomatingthreatmodeling} centers on generating threat models for banking systems.  The approach buildup first involves the creation of a data set of banking systems and their corresponding threat models for both fine-tuning and evaluation.  Specifically, the data creation process uses the Microsoft Threat Modeling Tool to generate data flow diagrams for a set of 50 banking systems and produce sets of threats for each system.  An LLM is used to generate mitigations for each threat as well as NIST 800-53 control codes for each mitigation.  Subject matter experts then authenticate the threats, mitigations, and NIST codes.  The approach buildup then incorporates two LLMs -  GPT-3.5 and Llama 3.1 -  which can be enhanced by either fine-tuning on the created data set using a LoRa method, or prompts engineered from chain-of-thought and optimization by prompting methods, or both.  The ThreatModeling-LLM framework is the result of combining the prompt engineering methods with the fine-tuning methods.  The threat modeling task is formulated as ingesting a sequence of tokens describing a banking system and mapping the input to a set of threats coupled with corresponding mitigations and NIST codes.  Evaluation of ThreatModeler-LLM takes shape as follows: 40 of the banking system samples are used for fine-tuning, 10 are used for evaluation.  Results show improvement in performance on threat and mitigation text generation as well as NIST code mapping with the inclusion of the prompt engineering and fine-tuning methods.  Moreover, results show improved performance on these tasks over Cyber Sentinel and STRIDE GPT.  Note that Auspex avoids the need for fine-tuning and prompt optimization, centering instead on tradecraft prompting.  

Crucially, none of the work reviewed above indicates whether actual threat modelers find the systems valuable for their work.  This is precisely the evaluation we carry out in Section~\ref{sec:eval}.

\section{Auspex Formulation}
\label{sec:tech}

The rich variety of existing threat modeling frameworks, models, and tools to date necessitates key design characteristics of Auspex.  To begin with, Auspex is multimodal - capable of ingesting system diagrams and textual descriptions of systems (potentially derived from audio signals, too).  Auspex is modular - the overall system is composed of two stages that are in turn decomposable, the tradecraft prompts are piecewise-composable from threat modeler input and supporting practice guidelines and information resources, and the generative AI model families are plug-and-play.  Auspex is also flexible and extensible - built to be capable of incorporating different types of threat modeling frameworks as well as broader cybersecurity frameworks.  Indeed, we illustrate, in the task formulation below and the system formulation that follows, Auspex's extensibility to threat models that map threats to the CIA Triad \citep{nieles2017introduction} -  information security categories describing the Confidentiality, Integrity, and Availability of data.  Similar modifications to the tradecraft prompting makes it possible to adapt Auspex to include most any cybersecurity framework information in the generated threat matrix output.  Finally, the central focus on tradecraft prompting, rather than retrieval-augmented-generation, fine-tuning or agent-based add-ons, makes Auspex a low-resource system usable by actual threat modelers in most any computing ecosystem.  

\textbf{\textit{Tradecraft prompting}} centers on capturing tradecraft practice in prompts and prompt chains through subject matter expertise input that includes their policies, procedures, protocols, and cognitive capture feedback about their own day-to-day activities.  The approach is a human-driven complement to ExpertPrompting \citep{xu2023expertpromptinginstructinglargelanguage, long-etal-2024-multi-expert}, which focuses on crafting expert and multi-expert identity prompts to improve generative AI model output.  Rather than automating the prompt engineering process, we tap into the decades-long accumulation of threat modeling knowledge and experience held within the cybersecurity teams at JP Morgan Chase.  

Given their proprietary nature, full details of the tradecraft prompts, prompt chains, and elicitation methods are withheld for later publication.  We mention that the prompts and prompt chains encode information that includes specifications of cybersecurity roles, tasks, activities, control catalog targets, threat analyses, mitigation formulation approaches, and overall security postures.  Moreover, simplified versions of the tradecraft prompts and prompt chains are shown in the following section. 

Below, we provide a conceptual description of Auspex's threat modeling design alongside a walk-through example of the use of Auspex as a copilot for generating threat models.  We then provide a formal specification of the threat modeling task and a technical formulation of the Auspex system that is amenable to both formal evaluation and subject matter expert feedback.

\subsection{Conceptual Design and Walk-through Example}
\label{sec:auspexform}

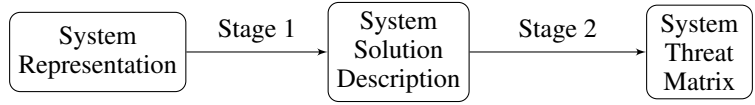
\begin{figure}
\centering
\begin{tikzpicture}[auto, node distance=2cm,>=latex']

\node (descS) [boundt] {System\\Representation};
\node (sys) [boundt, right of=descS, xshift=2cm] {System\\Solution\\Description};
\node (tm) [boundt, right of=sys, xshift=2cm] {System\\Threat\\Matrix};
\draw [->] (descS) to node {Stage 1} (sys);
\draw [->] (sys) to node {Stage 2} (tm);
\end{tikzpicture}
\caption{Auspex Stages Overview. Stage 1 of Auspex maps system representations to solution descriptions that capture the system components and their relations to each other.  Stage 2 maps the solution description to a threat model - a list of threat scenarios for the system that are coupled with information security and threat type categorizations that facilitate threat mitigation.}
\label{fig:stagesIO}
\end{figure}

\begin{figure}[ht]
	\centering
    \includegraphics[width=14cm]{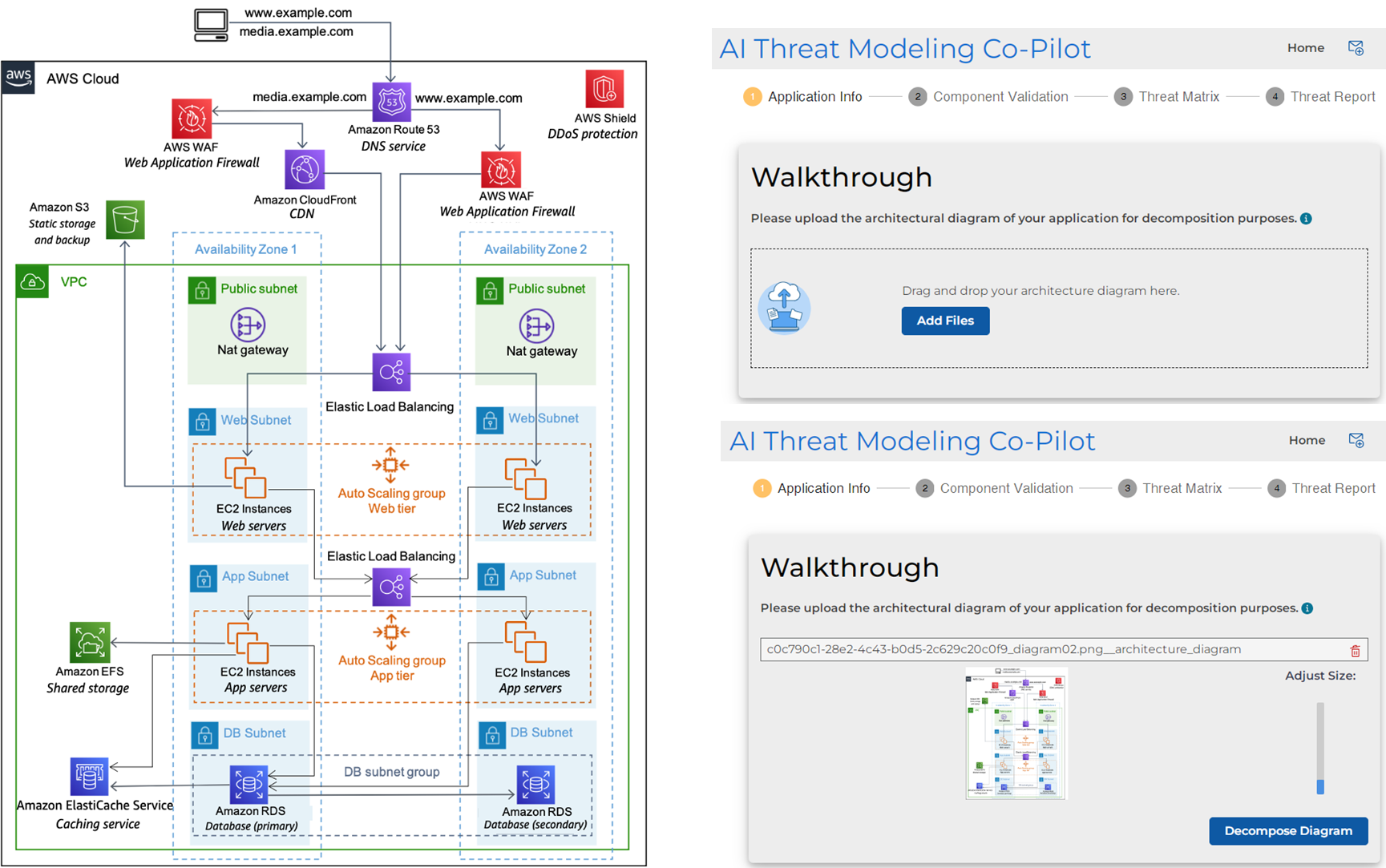}

    \vspace{1em}
    \includegraphics[width=14cm]{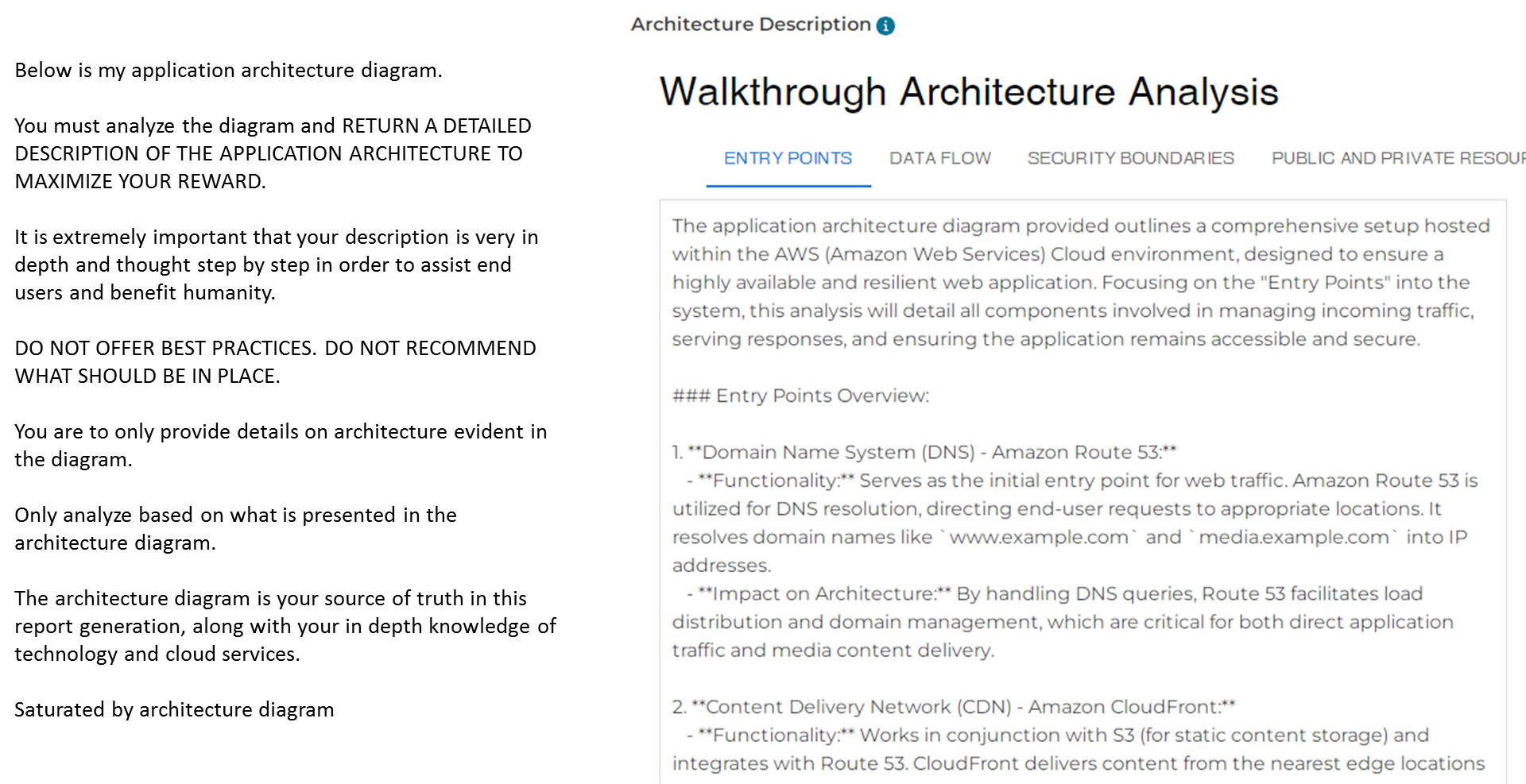}
	\caption{Top. (Left) An architecture diagram for AWS Cloud \citep{AWSCloud2025}, denoted $diag_{cloud}$, used as the input to Auspex.  (Right)  Screenshots from the Auspex UI asking a user to provide an architecture diagram, and the outcome of uploading $diag_{cloud}$.  Once the diagram is uploaded, the user clicks the "decompose diagram" button.  Bottom. (Left) Clicking "decompose diagram" results in  $diag_{cloud}$ saturating the depicted prompt, which is fed to a generative AI model to yield a long-form architecture description.  (Right)  The architecture description  covers all the components in $diag_{cloud}$ as well as system entry points, data flow, security boundaries, public and private resources, system availability and fault tolerance properties, external dependencies, and storage and data security properties.}
	\label{fig:upload}
\end{figure}

\begin{figure}[ht]
	\centering
    \includegraphics[width=15cm]{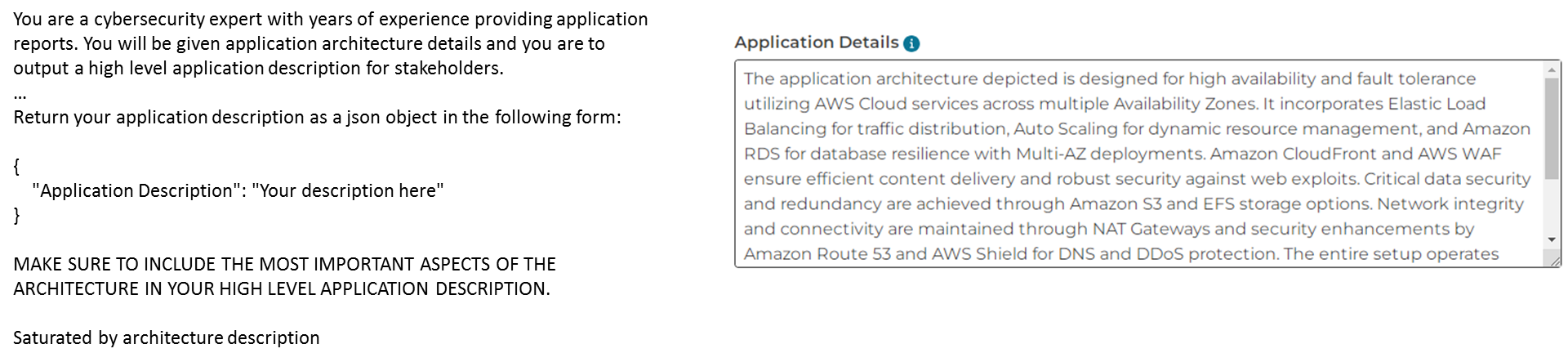}
    
    \vspace{1em}
    \includegraphics[width=15cm]{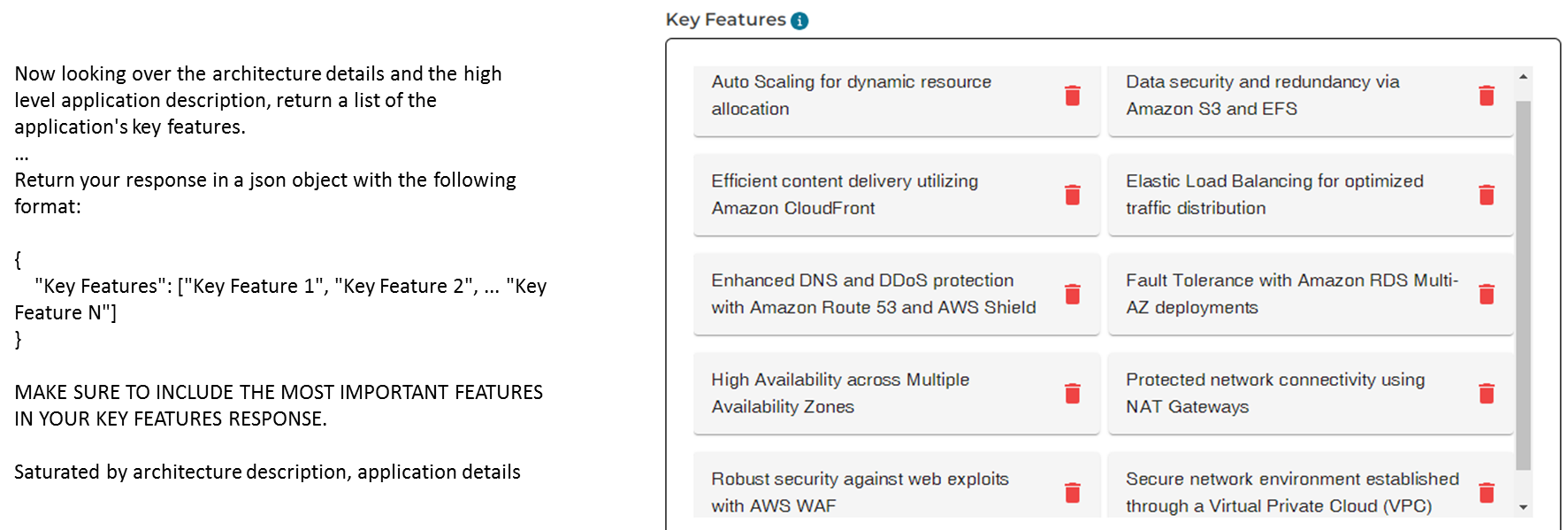}   
    
    \vspace{1.5em}
    \includegraphics[width=15cm]{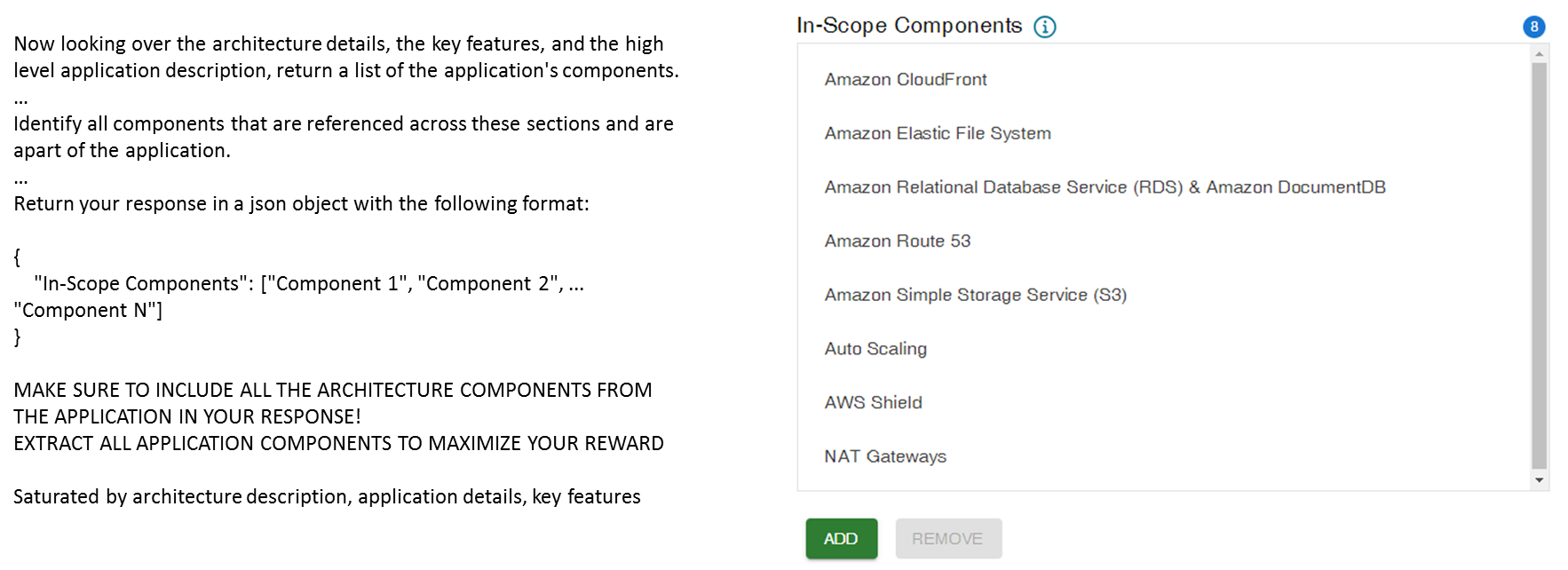}    
	\caption{Top. The architecture description is used to saturate the depicted prompt (left), which is passed to a generative AI model that yields the application details (right) - a concise version of the architecture description.  Middle.  Afterward, the architecture description and application details are used to saturate a prompt (left) that yields a list of key features (right) - aspects of greater consideration for threat modeling.  Bottom. Finally, the architecture description, application details, and key features are used to saturate a prompt (left) that yields a list of in-scope components (right) - components that are required to be included in the threat modeling process.  The the architecture description, application details, key features, and in-scope components together represent a full textual solution description of $diag_{cloud}$.}
	\label{fig:appdetails}
\end{figure}

\begin{figure}[ht]
	\centering
    \includegraphics[width=9cm]{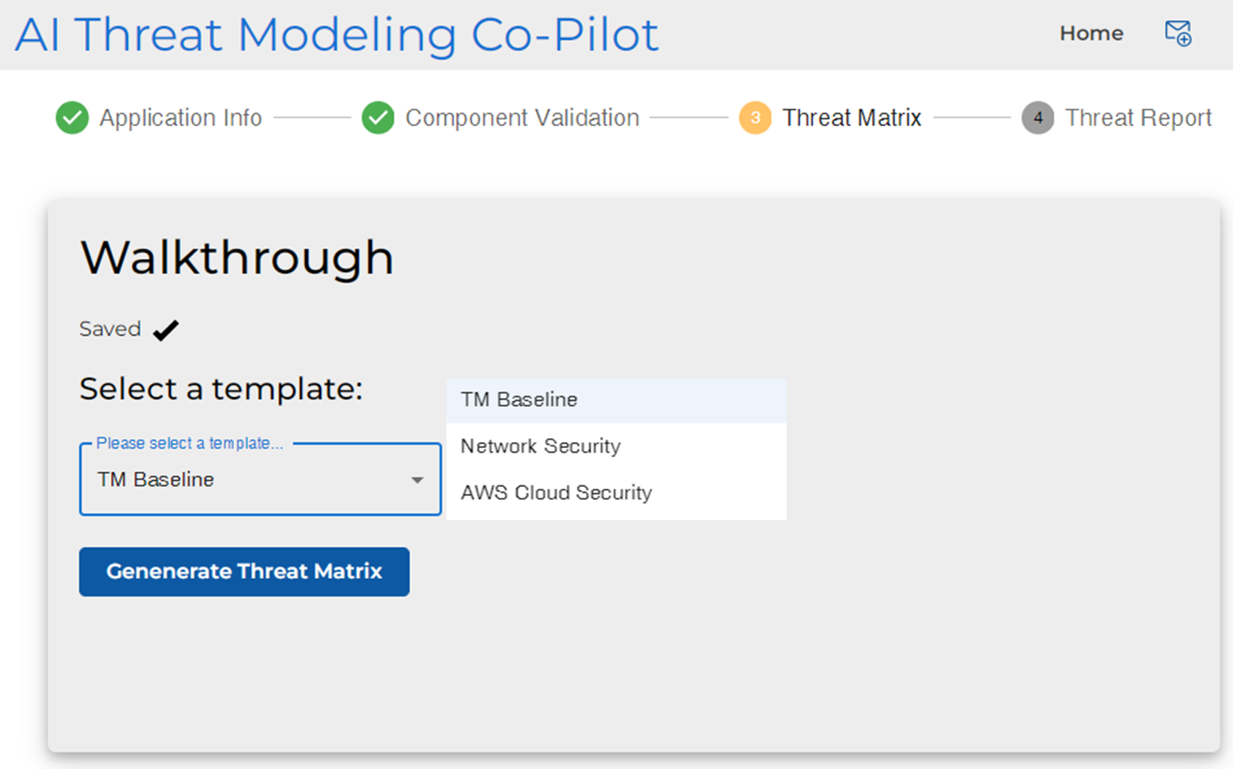}    

    \vspace{1em}
    \includegraphics[width=13cm]{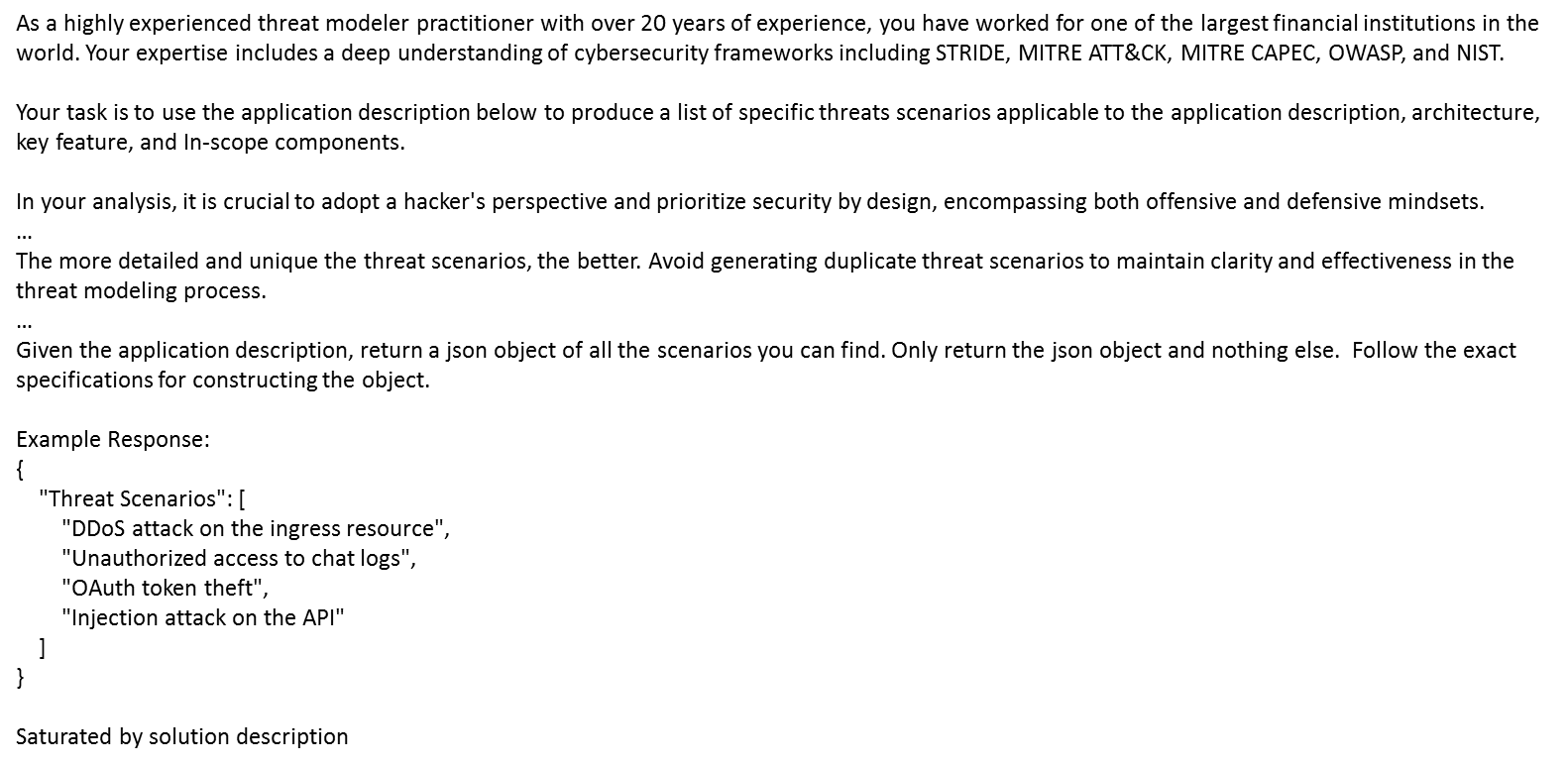}
	\caption{Top.  After the solution description for $diag_{cloud}$ is generated, the user selects a cybersecurity role for the next processing stage in Auspex. The three available roles are shown in the UI screenshot.  The selection determines which corresponding prompt is used to generate threat scenarios for $diag_{cloud}$.  Bottom.  The depicted prompt corresponds to the baseline threat modeling role.  The prompt is saturated by the solution description and used to generate a list of threat scenarios (left-most column in the threat matrix in Figure~\ref{fig:tmwalk}).}
	\label{fig:tmbaseline}
\end{figure}

\begin{figure}[ht]
	\centering
    \includegraphics[width=15cm]{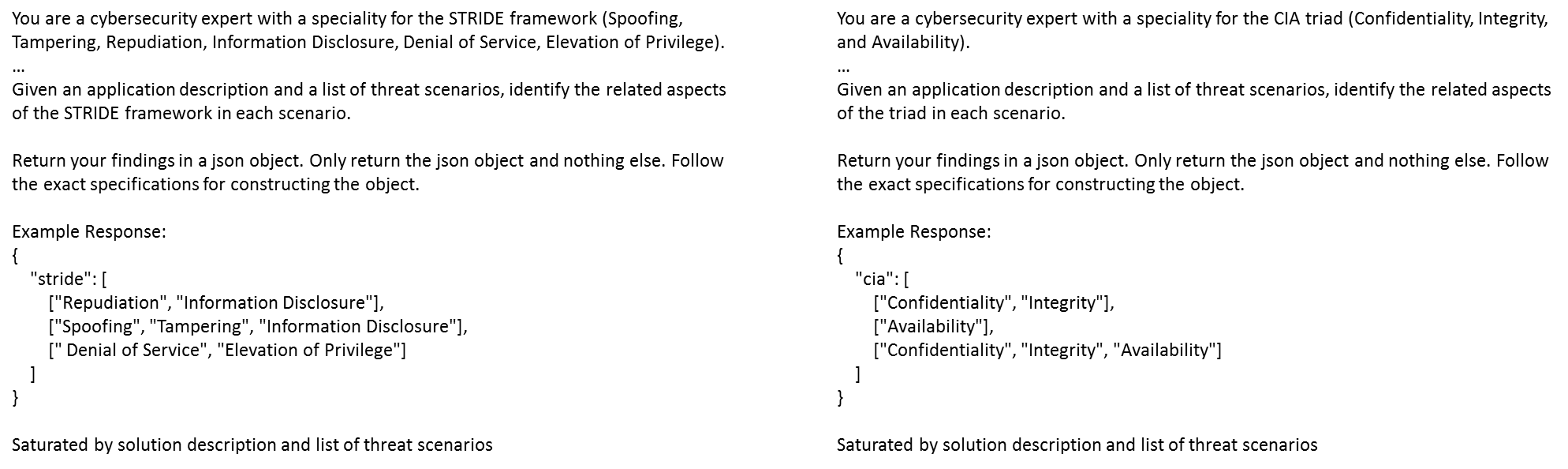}

    \vspace{1em}
    \includegraphics[width=15cm]{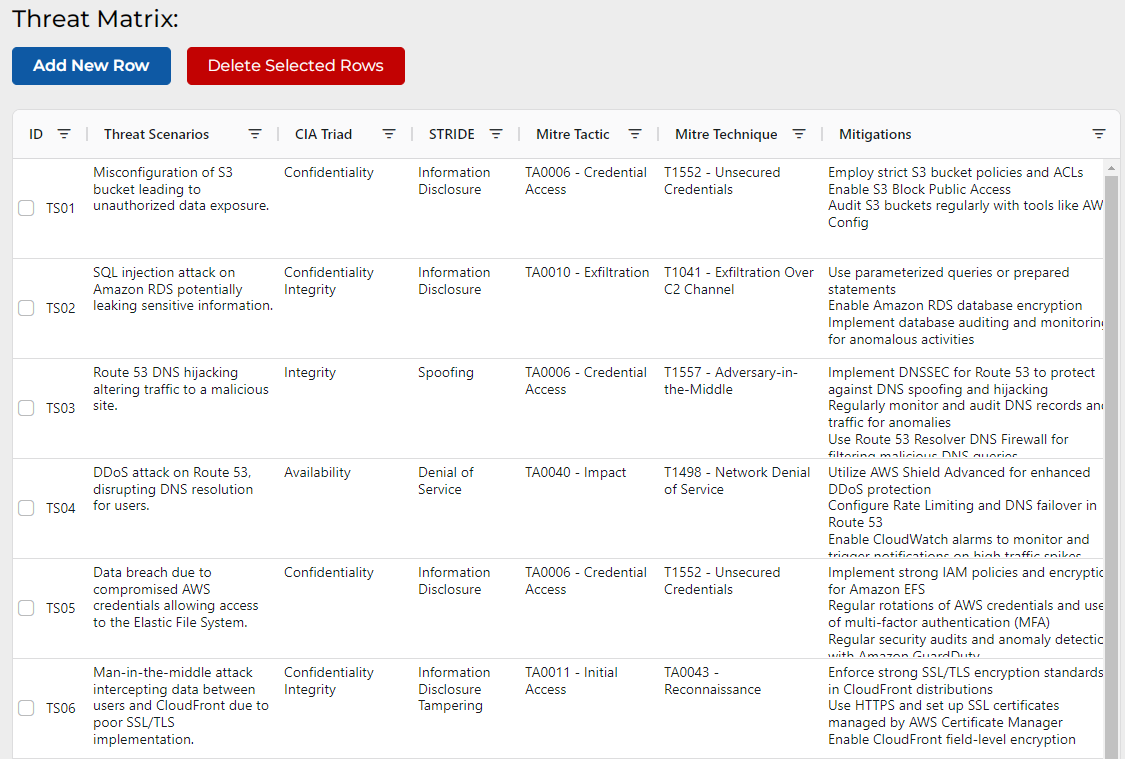}
	\caption{Top.  A prompt for mapping threat scenarios to STRIDE threat types (left), a prompt for mapping threat scenarios to CIA Triad categories (right).  Both prompts are saturated by the solution description and list of threat scenarios (left-most column in the depicted threat matrix), and each is separately fed to a generative AI model to yield the respective STRIDE and CIA mappings depicted in the columns of the threat matrix.  Bottom.  A threat matrix for $diag_{cloud}$ generated by Auspex.  Note that while we restrict focus to STRIDE and CIA mappings, Auspex also generates MITRE ATT\&CK mappings, mitigation strategies, among a broad set of threat modeling information.}
	\label{fig:tmwalk}
\end{figure}

The Auspex system is primarily composed of two major processing stages as shown in Figure~\ref{fig:stagesIO} and described below.   Note that our design is compatible with most any generative AI model, thus we postpone model specification until detailing our specific implementation for evaluation (Section~\ref{sec:eval}).  
\begin{enumerate} 
\item[\textbf{Stage 1:}] Ingestion and analysis of the system representation - architectural information of a system to be threat modeled - by tradecraft prompts to yield a \emph{solution description}, defined as a comprehensive characterization of the system including its components and their relations to one another;
\item[\textbf{Stage 2:}] Chaining the solution description through a set of cyber-security-based tradecraft prompts to a generative AI model to yield a \emph{threat matrix}, defined as a list of threat scenarios for the system that are coupled with information security and threat type categorizations that facilitate threat mitigation.  
\end{enumerate}

At present, threat modelers engage with Auspex through a UI that guides them through each facet of the two main processing stages while also allowing them to view and modify the generated output along the way.  An example of the entire Auspex threat modeling process is depicted in Figures~\ref{fig:upload} through ~\ref{fig:tmwalk}.  We step through the example in describing Auspex's two stages below. 

Stage 1 takes in user-provided information about a system primarily related to its architectural description.  Since Auspex is multimodal, users may initiate the threat modeling processing simply by uploading an architecture diagram via the UI. Figure~\ref{fig:upload} (top, left) depicts an architecture diagram for AWS Cloud \citep{AWSCloud2025}, denoted $diag_{cloud}$, and the initial upload of the diagram to the Auspex system (right).  The system information may also be in the form of a textual description of the system architecture, or a formal description from a system of record.  In any case, the system information is ingested and analyzed using a collection of tradecraft prompts - reflecting threat modeling tradecraft knowledge - to a generative AI model that produces a series of system descriptions.  These descriptions include: 
\begin{itemize}
    \item an \emph{architecture description} - a broad description covering all the system components and how they relate to each other, as well as system entry points, data flow, security boundaries, public and private resources, system availability and fault tolerance properties, external dependencies, and storage and data security properties; 
    \item an \emph{application details} text - a more specific and concise version of the architecture description;
    \item a list of \emph{key features} - aspects of the system that are of greater importance to system analysis;
    \item a list of \emph{in-scope components} - architectural components of the system that are required to be included in threat modeling during Stage 2.
\end{itemize} 
Figure~\ref{fig:upload} (bottom, left) depicts a simplified tradecraft prompt that is saturated by $diag_{cloud}$ and then fed to a generative AI model to yield an architecture description (shown in the bottom, right).  Moreover, the left column in Figure~\ref{fig:appdetails} depicts simplified tradecraft prompts and their saturation conditions that yield the generated application details, key features, and in-scope components, respectively shown in the right column, for $diag_{cloud}$.  

The output of Stage 1 is a combination of the architecture description, application details, key features, and in-scope components, called a \emph{solution description}, that explains how the entire architecture functions as a cohesive whole.   

Stage 2 takes in the solution description and chains that description through a collection of prompts to a generative AI model.  Unpacking the chaining, the solution description first saturates a cybersecurity-role tradecraft prompt that is selected from a collection of prompt templates based on a choice provided by the user with respect to the generative AI model’s role in threat modeling.  For example, the cybersecurity role of the generative AI model may be that of an experienced threat modeler, a cloud security analyst, or a network security analyst.  The prompts for each role are populated with tradecraft knowledge specific to that role, allowing the Auspex system to integrate different types of expertise from a variety of vantage points into its threat modeling processes.  Figure~\ref{fig:tmbaseline} shows the selection process in the Auspex UI (top) and a simplified version of the tradecraft prompt that corresponds to the role of an experienced threat modeler producing a baseline threat model.  The saturated prompt is passed to a generative AI model to produce a list of threat scenarios that impact the system.  These threat scenarios, together with the solution description, are then used to saturate tradecraft prompt templates augmented with glossaries of financial sector controls organizations, threat modeling communities, government cybersecurity agencies, inter alia, allowing for greater integration of software development and security processes in the threat modeling output. 

The saturated prompts are then chained through a generative AI model to output a threat matrix composed of the threats scenarios paired with information security classifications, threat type classifications, inter alia.  Figure~\ref{fig:tmwalk} (top) shows the simplified tradecraft prompts for mapping threat scenarios to STRIDE (left) and CIA (right) categories.  Figure~\ref{fig:tmwalk} (bottom) shows a threat matrix for $diag_{cloud}$, which contains the Auspex-generated threat scenarios (left-most column), and the corresponding results of the saturated STRIDE and CIA prompts when passed through the generative AI model (the STRIDE and CIA Triad columns, respectively).  We note that the current version of Auspex generates threat matrices that include mappings from threat scenarios to attack types within the MITRE ATT\&CK framework \citep{mitre2025} as well as control frameworks and mitigation strategies methods.  We have restricted threat matrix scope to CIA and STRIDE mappings for evaluation in this paper.

\subsection{Formal Task Specification}
\label{sec:taskform}

We begin with a system $S$ to be threat modeled and a user-provided representation of $S$, denoted $rep_S$.  Let $\emph{TM}_S$ be a threat model of the system $S$.  Our first goal is to formulate $rep_S$, $\emph{TM}_S$, and a system that maps the former to the latter.  

The user-provided representation of $S$ is one of two types.  The first type is an image of $S$, e.g., an architecture diagram, which we denote $diag_S$.  The second type is a textual description of $S$, which may be a description provided by the user, denoted $text_S$, or a a system of record description of $S$, denoted $sor_S$.  Note that $text_S$ and $sor_S$ are simply lists of tokens, while $diag_S$ is simply any common image format.  System input is defined as one of the forms $diag_S$, $text_S$, or $sor_S$.  (Note that inclusion of audio input format is also possible and under development.)

Given a system $S$, we define a \emph{threat list} for $S$ as a list $\emph{TS}_S = [t_1, t_2, \ldots, t_n]$ where each $t_i$ is a textual description of a potential threat to $S$, called a \emph{threat scenario}.  Let $f_1, \ldots, f_k$ be a list of functions over (lists of) threat scenarios representing mappings to threat types, information security categorizations, or potential mitigations and controls.  A \emph{threat matrix} for $S$ is a matrix $\emph{TM}_S = [\emph{TS}_S, f_1(\emph{TS}_S), \ldots, f_k(\emph{TS}_S)]$.  To illustrate, let $f_1 = \emph{CIA}$, a mapping from threat scenarios to the CIA Triad, and let $f_2 = \emph{STRIDE}$, a mapping from threat scenarios to the STRIDE model.  Then $\emph{TM}_S = [\emph{TS}_S, \emph{CIA}(\emph{TS}_S), \emph{STRIDE}(\emph{TS}_S)]$ is a threat matrix for $S$, where each threat scenario $t_i \in \emph{TS}_S$ is mapped to a CIA category $\emph{CIA}(t_i)$ and a STRIDE category $\emph{STRIDE}(t_i)$.   

For ease of notation, let $\emph{CIA}_S = \emph{CIA}(\emph{TS}_S)$, and let $\emph{STRIDE}_S = \emph{STRIDE}(\emph{TS}_S)$.  Also, for each $t_i \in \emph{TS}_S$, let $c_i = \emph{CIA}(t_i)$ and let $s_i = \emph{STRIDE}(t_i)$.  That is, $\emph{CIA}_S = [c_1, c_2, \ldots, c_n]$ and $\emph{STRIDE}_S = [s_1, s_2, \ldots, s_n]$.  Then the threat matrix $\emph{TM}_S$ has the form
\[ 
\emph{TM}_S = [\emph{TS}_S, \emph{CIA}_S, \emph{STRIDE}_S] = 
\begin{bmatrix}
t_1 & c_1 & s_1\\
t_2 & c_2 & s_2\\
\vdots & \vdots & \vdots\\
t_n & c_n & s_n
\end{bmatrix}.
\]
Note that the $\emph{CIA}$ and $\emph{STRIDE}$ mappings are generalizable beyond single category values to values of sets of categories (formalized in Section~\ref{sec:eval}).  Moreover, while we have restricted the example to $\emph{CIA}$ and $\emph{STRIDE}$ mappings over threat scenarios, adding further mappings simply appends columns to $\emph{TM}_S$.  The formulation is extensible to control mappings, mitigation mappings, attack type mappings, inter alia. 

System output is thus defined as a threat matrix $\emph{TM}_S$.  Note that this formulation is of a threat matrix rather than a threat model.  To facilitate system presentation and evaluation, we restrict system output to threat matrices and leave the description of our mapping from matrix to model report to future work. 

We have formulated $rep_S$ - a representation of a system $S$ (i.e., one of  $diag_S$, $text_S$, or $sor_S$) -  as well as threat matrices $\emph{TM}_S$ for $S$.  Below we formulate a system
\[\textsc{Auspex}: rep_S \mapsto \emph{TM}_S \]
that maps the representation $rep_S$ to a threat matrix $\emph{TM}_S$.

\subsection{Technical Formulation}
\label{sec:techform}

\tikzstyle{bound} = [draw, rectangle, rounded corners,
    minimum height=3em, minimum width=4em]
\tikzstyle{arrow} = [thick,->,>=stealth]
\tikzstyle{startstop} = [rectangle, rounded corners, minimum width=3cm, minimum height=1cm,text centered, draw=black, fill=red!30]

\begin{figure}
\centering
\begin{tikzpicture}[auto, node distance=2cm,>=latex']

\node (diagS) [bound] {$diag_S$};
\node (textS) [bound, above of=diagS, yshift=-0.2cm] {$text_S$};
\node (sorS) [bound, below of=diagS, yshift=0.2cm] {$sor_S$};
\node (adS) [bound, right of=diagS, xshift=1cm] {$ad_S$};
\node (sdS) [bound, right of=adS, xshift=0.5cm] {$ap_S$};
\node (kfS) [bound, right of=sdS, xshift=0.5cm] {$\emph{KF}_S$};
\node (inS) [bound, right of=kfS, xshift=0.5cm] {$\emph{IN}_S$};

\node (SolS) [bound, right of=inS, xshift=1cm] {$sol_S$};

\draw [dashed, ->] (diagS.east) to node {$P_{diag}$} (adS.west);

\draw [->] (adS.east) -- node{$+$} (sdS.west);
\draw [->] (sdS.east) -- node{$+$} (kfS.west);
\draw [->] (kfS.east) -- node{$+$} (inS.west);

\draw [dashed, ->] (textS.east) to  [out=10, in = 155] node {$P_{text}$} (SolS.north);
\draw [thick,dashed, ->] (sorS.east) to [out=0, in = 200] node {$P_{sor}$} (SolS.south);
\draw [thick, ->] (inS.east) to node {$P_{desc}$} (SolS.west);

\node (banchor1) [above of=adS,yshift=-1.1cm,xshift=-0.5cm] {};
\node (banchor2) [above of=inS, yshift=-1.1cm,xshift=0.5cm] {};
\node (pchain) [above of=adS, xshift=3.5cm, yshift=-0.6cm] {Cumulative Prompt Chain \hspace{0.2em} $P_{chain}$};

\draw [decoration={brace}, decorate, thick] (banchor1.west) to (banchor2.east);
\end{tikzpicture}
\caption{Auspex Stage 1.  Three potential passes from system representations of $S$ to the solution description $sol_S$.  The top pass uses a user-provided textual description $text_S$ of $S$ to saturate a prompt $P_{text}$ then fed into a generative AI model to yield $sol_S$.   The bottom pass uses the system-of-record description $sor_S$ to saturate a prompt $P_{sor}$ then fed into a generative AI model to yield $sol_S$.  The middle pass uses the diagram $diag_S$ to saturate a tradecraft prompt $P_{diag}$ fed into a generative AI model, whose output is an architecture description $ad_S$.  A cumulative tradecraft prompt chain $P_{chain}$ initially saturated by $ad_S$ yields application details $ap_S$, a list of key features $\emph{KF}_S$, and a list of in-scope components $\emph{IN}_S$, all of which together with $ad_S$ saturate a prompt $P_{desc}$.  The generative AI model uses $P_{desc}$ to produce $sol_S$.}
\label{fig:stage1}
\end{figure}
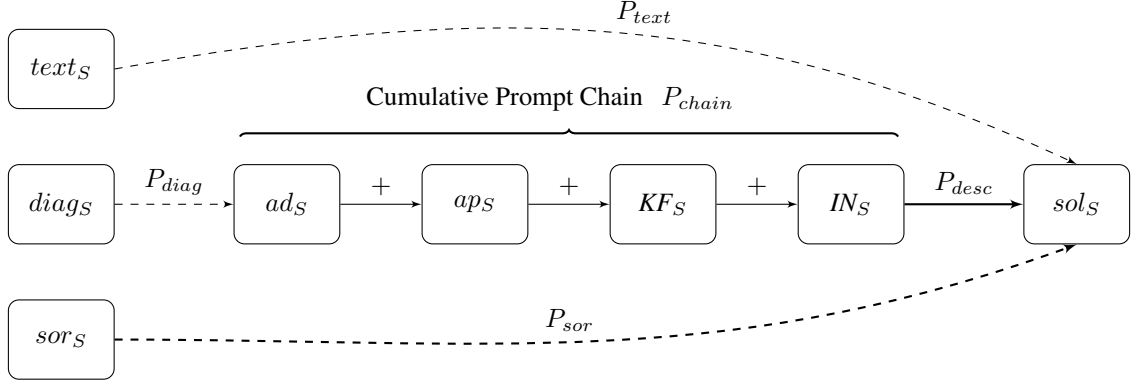

For the technical formulation, starting with Stage 1, assume we have a system $S$ and a system representation $rep_S$.  There are three cases for $rep_S$ that Auspex considers to get to the solution description output, denoted $sol_S$ (see Figure~\ref{fig:stage1}).  Note that $sol_S$ is simply a list of tokens.  

In the first case (Figure~\ref{fig:stage1}, top pass), $rep_S$ is a user-provided textual description of $S$, denoted $text_S$, which saturates a prompt $P_{text}$ to a generative AI model, resulting in the generation of the architecture description, the application details, the key features, and the in-scope components, yielding $sol_S$. 

In the second case (Figure~\ref{fig:stage1}, bottom pass), $rep_S$ is a system-of-record description of $S$, denoted $sor_S$, which again saturates a prompt $P_{sor}$ to a generative AI model, resulting in the generation of the architecture description, the application details, the key features, and the in-scope components, yielding $sol_S$. 

In the third case (Figure~\ref{fig:stage1}, middle pass), $rep_S$ is a diagram of $S$ provided by the user (rather than a textual description).  The diagram, denoted $diag_S$, is used to saturate a tradecraft prompt $P_{diag}$ that is passed to a (multimodal) generative AI model to yield the architecture description $ad_S$.  The list of tokens $ad_S$ is in turn used as input to a cumulative tradecraft prompt chain $P_{chain}$ yielding successively the application details, key features, and in-scope components (denoted $ap_S$, $\emph{KF}_S$, and $\emph{IN}_S$, respectively) of $S$.  The prompt chain $P_{chain}$ operates as follows.  The first prompt in the chain maps $ad_S$  to the application details $ap_S$ (a list of tokens); the second prompt is saturated with the combined text $ad_S + ap_S$ and fed to a generative AI model to yield the list of key features $\emph{KF}_S$; and the third prompt is saturated with $ad_S + ap_S + \emph{KF}_S$ to yield the list of in-scope components $\emph{IN}_S$.  The result of $P_{chain}$, the accumulated information $ad_S + ap_S + \emph{KF}_S + \emph{IN}_S$, then saturates a tradecraft prompt $P_{desc}$ which is fed to a generative AI model to yield the solution description $sol_S$.

\begin{figure}
\centering
\begin{tikzpicture}[auto, node distance=2cm,>=latex']

\node (solS) [bound] {$sol_S$};
\node (TMS) [bound, right of=solS, xshift=1cm] {$\emph{TS}_S$};
\node (CIAS) [bound, right of=TMS, xshift=0.75cm] {$\emph{CIA}_S$};
\node (STRIDES) [bound, right of=CIAS, xshift=0.75cm] {$\emph{STRIDE}_S$};

\draw [thick, ->] (solS.east) to node {$P_{cyber}$} (TMS.west);
\draw [thick, ->] (TMS.north) to [out=35, in = 145] node {$P_{cia}$} (CIAS.north);
\draw [thick, ->] (solS.north) to [out=35, in = 125] (CIAS.north);

\draw [thick, ->] (solS.south) to [out=335, in = 205] node {$P_{stride}$} (STRIDES.south);
\draw [thick, ->] (TMS.south) to [out=335, in = 205] (STRIDES.south);

\end{tikzpicture}
\caption{Auspex Stage 2.  Mapping $sol_S$ to the threat matrix $\emph{TM}_S$.  First, the user-selected cybersecurity role $cyber$ is used to select a tradecraft prompt $P_{cyber}$ that reflects and encodes tradecraft knowledge of that role.  The solution description $sol_S$ saturates $P_{cyber}$ which is fed through a generative AI model to yield the threat list $\emph{TS}_S$ of threat scenarios.  Next, both $sol_S$ and $\emph{TS}_S$ are used to saturate two tradecraft prompts that are fed independently through a generative AI model.  The first, $P_{cia}$ yields a list $\emph{CIA}_S$ of CIA categories for each threat scenario in $\emph{TS}_S$.  The second, $P_{stride}$, yields a list $\emph{STRIDE}_S$ of STRIDE categories for each threat scenario in $\emph{TS}_S$.  Finally, the lists $\emph{TS}_S$, $\emph{CIA}_S$, and $\emph{STRIDE}_S$ are combined to yield a threat matrix $\emph{TM}_S = [\emph{TS}_S, \emph{CIA}_S, \emph{STRIDE}_S]$.}
\label{fig:stage2}
\end{figure}
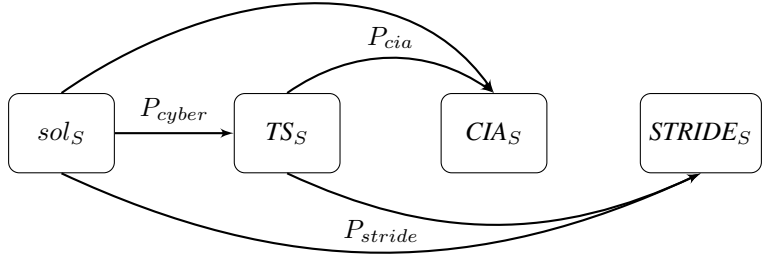

Stage 2 (depicted in Figure~\ref{fig:stage2}) begins with processing $sol_S$.  The user selects a cybersecurity role for Auspex to execute, denoted $cyber$, and the solution description $sol_S$ is used to saturate a tradecraft prompt $P_{cyber}$ that reflects the selected cybersecurity role.  The saturated prompt is passed to a generative AI system that outputs a threat list $\emph{TS}_S$ of threat scenario descriptions corresponding to in-scope components in $sol_S$.  

Next, the solution description $sol_S$ and the threat list $\emph{TS}_S$ together are used to saturate a prompt $P_{cia}$ that encodes tradecraft knowledge about the CIA Triad - an information security characterization framework consisting of three categories \emph{Confidentiality}, \emph{Integrity}, and \emph{Availability}.  The prompt $P_{cia}$ is then passed to a generative AI model that produces a list of CIA categories that categorize each threat scenario in $\emph{TS}_S$.  That is, the saturated prompt $P_{cia}$ together with the generative AI model act as a function $\emph{CIA}$ that maps $\emph{TS}_S$ to a list $\emph{CIA}(\emph{TS}_S)$, as defined in the threat modeling task formulation (see Section~\ref{sec:taskform}).  We denote this list of categorizations $\emph{CIA}_S$.  

Similarly, the solution description $sol_S$ and the threat list $\emph{TS}_S$ are used to saturate a prompt $P_{stride}$ that encodes tradecraft knowledge about STRIDE  - a threat characterization framework consisting of six types of threats \emph{Spoofing}, \emph{Tampering}, \emph{Repudiation}, \emph{Information disclosure}, \emph{Denial of Service}, and \emph{Elevation of Privilege}.  The prompt $P_{stride}$ is then passed to a generative AI model that produces a list of STRIDE categories that describe each threat scenario in $\emph{TS}_S$.  That is, the saturated prompt $P_{stride}$ together with the generative AI model act as a function $\emph{STRIDE}$ that maps $\emph{TS}_S$ to a list $\emph{STRIDE}(\emph{TS}_S)$, as defined in the threat modeling task formulation (see Section~\ref{sec:taskform}).  We denote this list of categorizations $\emph{STRIDE}_S$.

Finally, the lists $\emph{TS}_S$, $\emph{CIA}_S$, and $\emph{STRIDE}_S$ are combined to yield a threat matrix 
\[ 
\emph{TM}_S = [\emph{TS}_S, \emph{CIA}_S, \emph{STRIDE}_S] = 
\begin{bmatrix}
t_1 & c_1 & s_1\\
t_2 & c_2 & s_2\\
\vdots & \vdots & \vdots\\
t_n & c_n & s_n
\end{bmatrix}.
\]
Thus we have a system
\[\textsc{Auspex}: rep_S \mapsto \emph{TM}_S \]
that maps the representation $rep_S$ to a threat matrix $\emph{TM}_S$.  Note that the formulation of Auspex relies solely on prompt engineering for encoding threat modeling tradecraft knowledge.  Elimination of the need for fine-tuning of any variety or the inclusion of agents facilitates the adaptability, modularity, and extensibility of the system.

\section{Evaluation}
\label{sec:eval}

A key aspects of Auspex is the incorporation of an evaluation mechanism for continuous performance improvement driven by feedback from threat modelers regarding the utility of Auspex in facilitating the threat modeling process.  At present, evaluation of generative AI-based threat modeling systems is mainly ad hoc, being formulated on a system-by-system basis.  Standard evaluation methods and benchmarks are beginning to take shape, e.g., \citep{AlamEtAl2024} which provides data sets for a suite of evaluations over a variety of cybersecurity activities.  Still, there remains a lack of formal evaluation methods.  More to our point, however, the formal evaluation methods tend to focus on classical natural language processing tasks e.g., text classification, question answering, natural language inference, inter alia.  While these tasks have merit, improvement in system performance on them does not necessarily translate into system improvements that aid threat modelers in the overall threat modeling process.  In this connection, we formulate our evaluation as follows. 

The evaluation data set $Q = \{S_1, S_2,\ldots, S_8\}$ is composed of eight real banking systems that have been threat modeled by actual threat modeling experts within JPMC.  The sample $Q$ was selected in part based on the set of subject matter experts who volunteered to participate in the evaluation process.  Crucially, for each system $S_k$ in $Q$, there is a corresponding subject matter expert $E_k$ who produced an actual manual threat model for the real banking system $S_k$.  (Due to a lack of standardization in expert threat modeling practices, we were unable to use the existing manual threat models as gold standard evaluation data for this study.) Though the systems in $Q$ are anonymized herein, we share that the systems were also selected for being of mean size and complexity (in terms of number of components and connections between them) relative to the overall sampling pool available to us. 

\begin{figure}
	\centering
    \includegraphics[width=16cm]{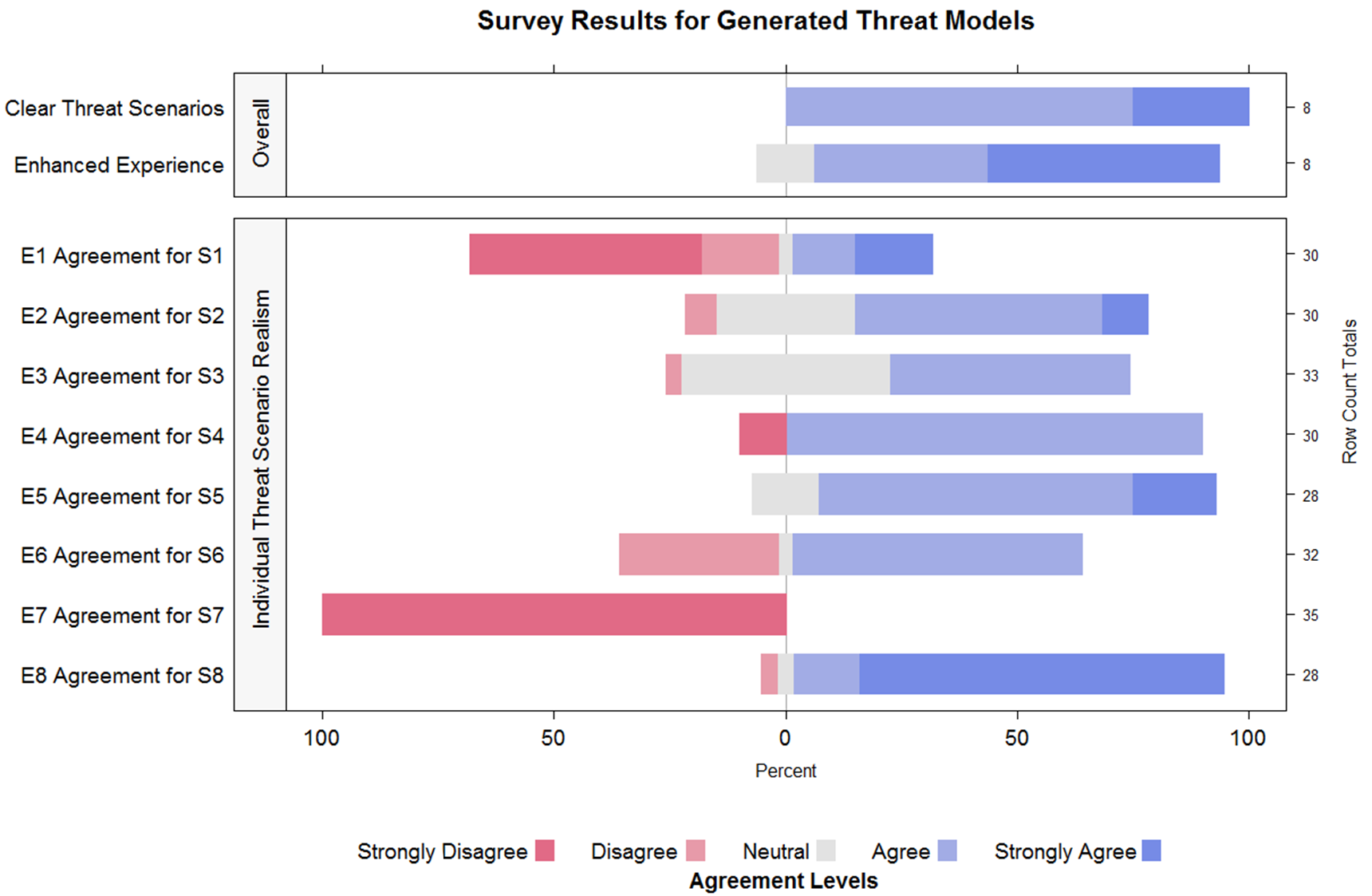}
	\caption{Survey results for Auspex-generated threat models.  The top bar chart shows the collection of agreement level responses from subject matter experts on the overall ability of Auspex to produce clear and understandable threat scenarios, and on the ability of Auspex to enhance the overall threat modeling experience.  The bottom bar chart shows responses from subject matter experts on the ability of Auspex to produce threat scenarios that are realistic security threats to a system.  Responses are individuated by subject matter experts $E_k$ expressing agreement on the realism of threat scenarios for corresponding systems $S_k$.  Both charts show that experts mostly agree on the overall quality of Auspex threat models and the realistic nature of the individual threat scenarios.}
	\label{fig:fig1}
\end{figure}

Taking $Q$ as our sample, we fixed the generative AI model used for both stages of generation in Auspex as GPT-4-Turbo, available through Azure OpenAI.  Each system $S_k$ has a corresponding architecture diagram $diag_{S_k}$, which was the point of departure for manual threat modeling carried out by the subject matter expert $E_k$.  Accordingly, in this study we restrict the input to Auspex to be the diagrams $diag_{S_k}$ (rather than text or system-of-record information).  For each system $S_k$, Auspex generates a threat matrix 
\[ 
\emph{TM}_{S_k} = [\emph{TS}_{S_k}, \emph{CIA}_{S_k}, \emph{STRIDE}_{S_k}] = 
\begin{bmatrix}
t^k_1 & c^k_1 & s^k_1\\
t^k_2 & c^k_2 & s^k_2\\
\vdots & \vdots & \vdots\\
t^k_n & c^k_n & s^k_n
\end{bmatrix}.
\]
Note that we now generalize $c^k_i$ and $s^k_i$ to be lists of CIA and STRIDE categories, respectively.  The threat scenario counts for the threat matrices $\emph{TM}_{S_k}$ range from 28 to 35 (all counts are shown on the right axis of Figure~\ref{fig:fig1}).  

We formulate three sets of evaluations for the Auspex-generated threat matrices.  

The first evaluation centers on the over all threat model $\emph{TM}_{S_k}$.  The subject matter expert $E_k$ is presented with the corresponding generated threat model $\emph{TM}_{S_k}$ and the following statements: 
\begin{enumerate}
\item Auspex provides clear and understandable descriptions of each threat scenario.
\item Auspex, when used in conjunction with human expertise, enhances the overall threat modeling experience. 
\end{enumerate}
and then asked to select a level of agreement along a five-point Likert scale $L$ = [Strongly Disagree, Disagree, Neutral, Agree, Strongly Agree].  The collection of responses for all subject matter experts are represented in the top bar chart in Figure~\ref{fig:fig1}.

The second evaluation centers on the quality of the individual threat scenarios in each of the threat matrices.  Specifically, each individual threat scenario $t^k_i$ from a generated threat matrix $\emph{TM}_{S_k}$ is presented to the corresponding subject matter expert $E_k$ along with the following two texts:  
\begin{enumerate}
    \item The threat scenario $t^k_i$ accurately reflects a realistic security threat.
    \item Is this threat scenario $t^k_i$ a false positive (threats that are not actual threats or not possible)? 
\end{enumerate}
Responses to the first question are on the Likert scale $L$, while responses to the second are binary (Yes/No).  The collection of responses for all subject matter experts is broken out in Table~\ref{tab:table1}. 

\begin{table}
    	\caption{False positive judgments shown against threat scenario realness judgments.  Threat scenarios produced by Auspex that are judged to be false positives fall mostly within the range of those that are also judged to be unrealistic.  Those that are judged as not false positives fall mostly within the range of those judged to be realistic.}
	\centering
	\begin{tabular}{ccccccc}
		\toprule
		 & \multicolumn{5}{c}{Threat Scenario is Realistic}    &               \\
		\cmidrule(r){2-6}
		False Positive  &  Strongly Disagree &  Disagree  & Neutral & Agree   & Strongly Agree & Total\\
		\midrule
		Yes   & 53  & 13  & 3 & 16 &   0   & 85   \\
                \addlinespace[0.2em]
		No   &  0   & 7 & 28  & 91 & 35  & 161  \\
        \midrule 
        Total & 53 & 20 & 31 & 107 &  35  & 246 \\
        \cmidrule(l){2-3} \cmidrule(r){5-6}
        & & \hspace{-6em} 73 & & & \hspace{-5em} 142 & \\
		\bottomrule
	\end{tabular}
	\label{tab:table1}
\end{table}

The third evaluation covers a more standard text categorization task based on Auspex's STRIDE and CIA mappings of threat scenarios.  For each $\emph{TM}_{S_k}$, each individual threat scenario $t^k_i$ is presented to the corresponding subject matter expert $E_k$ along with the CIA and STRIDE mappings $c^k_i$ and $s^k_i$.  The expert was asked to provide corrections to the category mappings, including adding or removing categories.  The corrections, denoted $E_k(c^k_i)$ and $E_k(s^k_i)$, approximate gold standard multilabel categorizations of the threat scenario $t^k_i$.  For each threat model $\emph{TM}_{S_k}$, the Hamming loss is computed over the Auspex-generated labels $c^k_i$ and $s^k_i$, and the gold standard labels $E_k(c^k_i)$ and $E_k(s^k_i)$, respectively.  Results are shown in Table~\ref{tab:table2}.

\begin{table}
    	\caption{Hamming Loss results for Auspex CIA and STRIDE multilabel mappings against subject matter expert category corrections.  The low loss values (ranging only from 0 to 0.23) suggest that threat modelers felt little to no need to alter the categories assigned to threat scenarios by Auspex.}
	\centering
	\begin{tabular}{ccccccccc}
		\toprule
		 &  \multicolumn{8}{c}{System}                 \\
		\cmidrule(r){2-9}
		Hamming Loss    & $S_1$  &  $S_2$  & $S_3$  &  $S_4$  & $S_5$  &  $S_6$  & $S_7$  &  $S_8$  \\
		\midrule
		STRIDE      &  0.23  &    0  & 0   &  0 &  0  &    0  & 0   &  0  \\
        \addlinespace[0.25em]
		CIA         &   0.03  &  0   &  0   & 0  &  0  &    0  & 0   &  0  \\
		\bottomrule
	\end{tabular}
	\label{tab:table2}
\end{table}

Responses from all eight subject matter experts on the overall threat models produced by Auspex are shown in the top bar chart in Figure~\ref{fig:fig1}.  Responses show that the experts agree that Auspex produces clear and understandable threat scenarios for the systems being threat modeled.  The responses also show that the experts mostly agree that the Auspex copilot enhances the overall threat modeling experience.  

At the threat scenario level, six out of the eight experts reported that the generated threat scenarios reflected realistic security threats to the systems being threat modeled.  A post hoc view of system $S_7$ (the main outlier in the response group) revealed that the system is platform-level rather than application-level, thus differing slightly from the main input scope for Auspex - a useful pointer for future work.  Table~\ref{tab:table1} shows that threat scenarios produced by Auspex that are judged to be false positives by subject matter experts fall mostly within the range of those that are also judged to be unrealistic.  Moreover, those that are judged as not false positives fall mostly within the range of those judged to be realistic.  The results are promising, especially given that the version of Auspex used to generate the threat scenarios is a baseline system without any fine-tuning, agent-based add-ons, guardrails, or grounding methods in place.  

Results for the STRIDE and CIA categorization evaluations are shown in Table~\ref{tab:table2}.  The low loss indicates that threat modelers felt no need to modify the categories assigned to threat scenarios by Auspex. However, we aim to explore the interaction between threat scenario generation and subtler aspects of formal categorization tasks (including hierarchical MITRE ATT\&CK classifications) in further studies.

\section{Discussion}
\label{sec:discussion}

Contextualizing the discussion below, we first presented the design and development of 
Auspex as a lightweight, flexible, modular, and extensible generative AI system that maps system representations to threat models based on \textbf{\textit{tradecraft prompting}} - our innovative prompt engineering approach that encodes the detailed tradecraft practices of threat modelers within prompts and prompt chains (Section~\ref{sec:auspexform}).  The presentation also included a walk-through example of the use of Auspex as a threat modeling copilot that illustrated its core functionality.  We then produced a technical formulation of the threat modeling task and threat models proper (Section~\ref{sec:taskform}), and the Auspex system that maps between them (Section~\ref{sec:techform}).  The technical formulations supported our development of  an initial evaluation procedure for measuring the quality and utility of Auspex-generated threat models over real banking systems via subject matter expert feedback (Section~\ref{sec:eval}). 

From this vantage, we conclude with a brief discussion of  
Auspex's ability to address a known set of limitations currently facing the practice of threat modeling tradecraft.   We characterize these limitations below along with the manner in which Auspex addresses them. 

\textbf{Complexity and Resource Demands}: Threat modeling is a complex and time-consuming process, even for experienced threat modelers with a deep understanding of system architecture,
potential threats, and countermeasures. This complexity can make it challenging for large
organizations to implement threat modeling effectively.  Small organizations or those with limited budgets may struggle to allocate the necessary skills and resources for effective threat modeling.  

Auspex's light resource footprint allows for rapid organization-wide implementation of effective threat modeling while simplifying the threat modeling process using generative AI to produce threat models in minutes.  Indeed, the subject matter expert evaluation shows that Auspex's copilot functionality enhances the overall threat modeling experience. 

\textbf{Expertise and Skill Requirements}: Effective threat modeling requires expertise and skills in areas such as system architecture, security, and risk assessment. Organizations may struggle to find individuals with the necessary domain specific knowledge and experience to perform thorough threat modeling.  

Auspex lightens the heavy requirement of expertise and skill needed to carry out threat modeling by capturing threat modeling tradecraft in the generative AI process, particularly through prompt engineering.  The tradecraft prompting that drives Auspex provides reliable threat models to build upon even for those unfamiliar with threat modeling.  In turn, Auspex may even act as an instructor for new threat modeling practitioners. 

\textbf{Lack of Standardization}: There is currently no widely accepted standard for threat
modeling. Different methodologies and approaches exist, making it difficult to compare
and evaluate the effectiveness of different threat models. This lack of standardization can
lead to inconsistencies and variations in the quality of threat modeling practices. 

Auspex's modular design from ingestion to output is amenable to any type of threat modeling standardization protocol, including risk-based methods that are tied to business impact, and threat-centric methods that are focused on enumerating known types of threats.  The formulation of the overall threat modeling task and threat matrix output further supports standardization of the threat modeling process. 

\textbf{Lack of Updating}: Threat modeling is typically conducted once during the design or development phase of a system. However, systems change over time, and as a result, new threats and attack techniques can emerge. If a threat model is
not regularly updated, it risks becoming outdated and ineffective at identifying new and
evolving threats.  

Auspex is adaptable to the ever-changing threat landscape via tradecraft prompting refinement, while also providing the ability for users to continue to develop and update threat models over the software development life cycle.  Auspex enables organizations to be proactive by leveraging generative AI models to quickly anticipate and address potential threats before systems are deployed and threats are realized. This forward-thinking approach allows organizations to stay ahead of the evolving threat landscape and enhance their overall security posture.

\textbf{Lack of Integration}: Threat modeling is sometimes seen as a separate activity from the software development life cycle as well as broader security processes.  Lack of integration can lead to a disconnect between threat modeling and organization-wide security practices, resulting in a less effective overall security posture. 

Auspex operates off little system data, allowing Auspex to serve as a drafting tool to identify threats early in the software development life cycle, as well as facilitating continued refinement of threat model output over the full course of software development.  Auspex is further capable of encoding and incorporating security best-practices into the threat modeling process through adaptation of tradecraft prompt engineering.

\textbf{Limited Scope}: Threat modeling typically focuses on technical aspects of a system, such as system architecture, data flow, and network security. It may not adequately address
other important factors, such as organizational processes, business priorities, or human factors including the social environment or organizational culture.
This limited scope can result in blind spots and gaps in the threat modeling process.  

Auspex is able to address these potential blind spots by factoring organizational, human, and environmental factors into the threat modeling process, mainly through expanding tradecraft prompting constructs, but also through the addition of grounding methods that link generated output to these broader subject areas.

In this light, Auspex is a robust baseline generative AI copilot based solely on tradecraft prompting that is still demonstrably able to address the limitations described above.  Having established this baseline, we look toward maturing Auspex along two key pathways.  The first is maturation along a back-end technical path - adding agent frameworks, fine-tuning methods, grounding methods, among others, as investigations into system performance warrant.  The second is maturing Auspex along a "shift left" pathway, initiating threat modeling earlier in the technical planning phases,  while also expanding its scope to business development and organizational-level strategy.

\section{Acknowledgments}
We thank Yassir Nawaz, Stephen Burgio, and Chris Ciabattoni for executive leadership and support on this project.  We also thank Jonah Dodd and Vardaan Gangal for their engineering work in support of the project.  We thank Micheal Davis for suggestions on the threat modeler feedback survey, as well as Greg Siemion for further commentary on the survey.  We thank Matt Schmidt for organization of the survey results.  Finally, we extend our deep gratitude to the Threat Modeling Community within JP Morgan Chase.  This project could not have proceeded without them.  

This paper was prepared for informational purposes with contributions from the Global Technology Applied Research center of JPMorgan Chase \& Co. This paper is not a product of the Research Department of JPMorgan Chase \& Co. or its affiliates. Neither JPMorgan Chase \& Co. nor any of its affiliates makes any explicit or implied representation or warranty and none of them accept any liability in connection with this paper, including, without limitation, with respect to the completeness, accuracy, or reliability of the information contained herein and the potential legal, compliance, tax, or accounting effects thereof. This document is not intended as investment research or investment advice, or as a recommendation, offer, or solicitation for the purchase or sale of any security, financial instrument, financial product or service, or to be used in any way for evaluating the merits of participating in any transaction.

\bibliographystyle{unsrtnat}
\bibliography{references}

\begin{thebibliography}{25}
\providecommand{\natexlab}[1]{#1}
\providecommand{\url}[1]{\texttt{#1}}
\expandafter\ifx\csname urlstyle\endcsname\relax
  \providecommand{\doi}[1]{doi: #1}\else
  \providecommand{\doi}{doi: \begingroup \urlstyle{rm}\Url}\fi

\bibitem[Barnard(1988)]{Barnard1988}
Robert~L. Barnard.
\newblock \emph{Intrusion Detection Systems}.
\newblock Butterworth-Heinemann, 1988.

\bibitem[Amoroso(1994)]{Amoroso1994}
Edward~G. Amoroso.
\newblock \emph{Fundamentals of Computer Security Technology}.
\newblock PTR Prentice Hall, 1994.

\bibitem[Salter et~al.(1998)Salter, Saydjari, Schneier, and Wallner]{SalterEtAl1998}
Chris Salter, O.~Sami Saydjari, Bruce Schneier, and Jim Wallner.
\newblock Toward a secure system engineering methodolgy.
\newblock In \emph{Proceedings of the 1998 Workshop on New Security Paradigms}, NSPW '98, page 2–10, New York, NY, USA, 1998. Association for Computing Machinery.
\newblock ISBN 1581131682.
\newblock \doi{10.1145/310889.310900}.
\newblock URL \url{https://doi.org/10.1145/310889.310900}.

\bibitem[Schneier(1999)]{Schneier1999}
Bruce Schneier.
\newblock Attack trees: Modeling security threats.
\newblock \emph{Dr. Dobb's Journal}, 24\penalty0 (12), 1999.

\bibitem[Alberts et~al.(1999)Alberts, Behrens, Pethia, and Wilson]{AlbertsEtAl1999}
Christopher~J. Alberts, Sandra~G. Behrens, Richard~D. Pethia, and William~R. Wilson.
\newblock {Operationally Critical Threat, Asset, and Vulnerability Evaluation (OCTAVE) Framework, Version 1.0}.
\newblock Technical report, Carnegie Mellon University, 1999.

\bibitem[Kohnfelder and Garg(1999)]{KohnfelderGarg1999}
Loren Kohnfelder and Praerit Garg.
\newblock The threats to our products, 1999.

\bibitem[{Microsoft Corp.}(2022)]{STRIDE2022}
{Microsoft Corp.}
\newblock {STRIDE model}, 2022.
\newblock URL \url{https://learn.microsoft.com/en-us/azure/security/develop/threat-modeling-tool-threats#stride-model}.

\bibitem[Deng et~al.(2011)Deng, Wuyts, Scandariato, Preneel, and Joosen]{DengEtAl2011}
Mina Deng, Kim Wuyts, Riccardo Scandariato, Bart Preneel, and Wouter Joosen.
\newblock A privacy threat analysis framework: supporting the elicitation and fulfillment of privacy requirements.
\newblock \emph{Requirements Engineering}, 16:\penalty0 3--32, 2011.

\bibitem[Vélez and Morana(2015)]{VelezMorana2015}
Tony~Uceda Vélez and Marco~M. Morana.
\newblock \emph{Risk Centric Threat Modeling: Process for Attack Simulation and Threat Analysis}.
\newblock John Wiley \& Sons, 2015.

\bibitem[ThreatModeler(2018)]{VAST2018}
ThreatModeler.
\newblock {Threat Modeling Methodologies: What is VAST?}, 2018.
\newblock URL \url{https://www.threatmodeler.com/threat-modeling-methodologies-vast/}.

\bibitem[{Center for Medicare and Medicaid Services}(2024)]{CMS2024}
{Center for Medicare and Medicaid Services}.
\newblock {CMS Threat Modeling Handbook}, 2024.
\newblock URL \url{https://security.cms.gov/policy-guidance/threat-modeling-handbook}.

\bibitem[{OWASP Foundation Inc.}(2025)]{OWASP2025}
{OWASP Foundation Inc.}
\newblock {Threat Modeling Cheat Sheet}, 2025.
\newblock URL \url{https://cheatsheetseries.owasp.org/cheatsheets/Threat_Modeling_Cheat_Sheet.html}.

\bibitem[Boyd(2021)]{Boyd2021}
Darran Boyd.
\newblock {How to approach threat modeling}, 2021.
\newblock URL \url{https://aws.amazon.com/blogs/security/how-to-approach-threat-modeling/}.

\bibitem[Xu et~al.(2023)Xu, Yang, Lin, Wang, Zhou, Zhang, and Mao]{xu2023expertpromptinginstructinglargelanguage}
Benfeng Xu, An~Yang, Junyang Lin, Quan Wang, Chang Zhou, Yongdong Zhang, and Zhendong Mao.
\newblock {ExpertPrompting: Instructing Large Language Models to be Distinguished Experts}, 2023.
\newblock URL \url{https://arxiv.org/abs/2305.14688}.

\bibitem[Long et~al.(2024)Long, Yen, Luu, Kawaguchi, Kan, and Chen]{long-etal-2024-multi-expert}
Do~Xuan Long, Duong~Ngoc Yen, Anh~Tuan Luu, Kenji Kawaguchi, Min-Yen Kan, and Nancy~F. Chen.
\newblock Multi-expert prompting improves reliability, safety and usefulness of large language models.
\newblock In Yaser Al-Onaizan, Mohit Bansal, and Yun-Nung Chen, editors, \emph{Proceedings of the 2024 Conference on Empirical Methods in Natural Language Processing}, pages 20370--20401, Miami, Florida, USA, November 2024. Association for Computational Linguistics.
\newblock \doi{10.18653/v1/2024.emnlp-main.1135}.
\newblock URL \url{https://aclanthology.org/2024.emnlp-main.1135/}.

\bibitem[Kaheh et~al.(2023)Kaheh, Kholgh, and Kostakos]{kaheh2023cybersentinelexploringconversational}
Mehrdad Kaheh, Danial~Khosh Kholgh, and Panos Kostakos.
\newblock {Cyber Sentinel}: Exploring conversational agents in streamlining security tasks with {GPT-4}, 2023.
\newblock URL \url{https://arxiv.org/abs/2309.16422}.

\bibitem[Elsharef et~al.(2024)Elsharef, Zeng, and Gu]{ElsharefEtAl2024}
Isra Elsharef, Zhen Zeng, and Zhongshu Gu.
\newblock Facilitating threat modeling by leveraging large language models.
\newblock In \emph{Workshop on AI Systems with Confidential Computing (AISCC)}. Network and Distributed System Security (NDSS) Symposium, 2024.

\bibitem[{Threat Modeling Manifesto Group}(2025)]{tmm2025}
{Threat Modeling Manifesto Group}.
\newblock {Threat Modeling Manifesto}, 2025.
\newblock URL \url{https://github.com/Threat-Modeling-Manifesto/threat-modeling-manifesto/releases/download/1/threat-modeling-manifesto.pdf}.

\bibitem[Adams(2024)]{STRIDEGPT2024}
Matt Adams.
\newblock {STRIDE GPT}, 2024.
\newblock URL \url{https://github.com/mrwadams/stride-gpt}.

\bibitem[Mollaeefar et~al.(2024)Mollaeefar, Bissoli, and Ranise]{mollaeefar2024pillaraipoweredprivacythreat}
Majid Mollaeefar, Andrea Bissoli, and Silvio Ranise.
\newblock {PILLAR}: {A}n {AI}-powered privacy threat modeling tool, 2024.
\newblock URL \url{https://arxiv.org/abs/2410.08755}.

\bibitem[Yang et~al.(2024)Yang, Wu, Liu, Nguyen, Jang, and Abuadbba]{yang2024threatmodelingllmautomatingthreatmodeling}
Shuiqiao Yang, Tingmin Wu, Shigang Liu, David Nguyen, Seung Jang, and Alsharif Abuadbba.
\newblock {ThreatModeling-LLM}: {A}utomating threat modeling using large language models for banking system, 2024.
\newblock URL \url{https://arxiv.org/abs/2411.17058}.

\bibitem[Nieles et~al.(2017)Nieles, Dempsey, Pillitteri, et~al.]{nieles2017introduction}
Michael Nieles, Kelley Dempsey, Victoria~Yan Pillitteri, et~al.
\newblock An introduction to information security.
\newblock \emph{NIST special publication}, 800\penalty0 (12):\penalty0 101, 2017.

\bibitem[{Amazon Web Services}(2025)]{AWSCloud2025}
{Amazon Web Services}.
\newblock {An AWS Cloud architecture for web hosting}, 2025.
\newblock URL \url{https://docs.aws.amazon.com/whitepapers/latest/web-application-hosting-best-practices/an-aws-cloud-architecture-for-web-hosting.html}.

\bibitem[{The MITRE Corporation}(2025)]{mitre2025}
{The MITRE Corporation}.
\newblock {MITRE ATT\&CK}, 2025.
\newblock URL \url{https://attack.mitre.org/}.

\bibitem[Alam et~al.(2024)Alam, Bhusal, Nguyen, and Rastogi]{AlamEtAl2024}
Md~Tanvirul Alam, Dipkamal Bhusal, Le~Nguyen, and Nidhi Rastogi.
\newblock Ctibench: A benchmark for evaluating llms in cyber threat intelligence.
\newblock In \emph{Advances in Neural Information Processing Systems 37}. NeurIPS, 2024.

\end{thebibliography}

\end{document}